\documentclass[runningheads]{llncs}
\usepackage{graphicx}
\usepackage{cite}
\usepackage{amsmath,amssymb,mathtools, amsfonts}
\usepackage{algorithmic}
\usepackage{algorithm2e}
\usepackage{textcomp}
\usepackage{subfig}
\usepackage{cleveref}
\usepackage{url}

\DeclareMathOperator{\argmax}{argmax}
\DeclareMathOperator{\argmin}{argmin}

\DeclareMathOperator{\conf}{conf}

\DeclareMathOperator{\sign}{sign}

\DeclareMathOperator{\D}{\mathcal{D}}

\DeclareMathOperator{\R}{\mathbb{R}}

\DeclareMathOperator{\X}{\mathcal{X}}
\DeclareMathOperator{\Y}{\mathcal{Y}}

\renewcommand{\vec}[1]{\mathbf{#1}}

\begin{document}
\title{Label Sanitization against Label Flipping Poisoning Attacks}
%
%
\author{Andrea Paudice \and
Luis Mu\~{n}oz-Gonz\'{a}lez \and
Emil C. Lupu}
\authorrunning{A. Paudice et al.}
%
\institute{Imperial College London, Department of Computing, London, UK \\
\email{a.paudice15@imperial.ac.uk \\l.munoz@imperial.ac.uk \\ e.c.lupu@imperial.ac.uk}}
\maketitle              
\begin{abstract}
Many machine learning systems rely on data collected in the wild from untrusted sources, exposing the learning algorithms to data poisoning. Attackers can inject malicious data in the training dataset to subvert the learning process, compromising the performance of the algorithm producing errors in a targeted or an indiscriminate way. Label flipping attacks are a special case of data poisoning, where the attacker can control the labels assigned to a fraction of the training points. Even if the capabilities of the attacker are constrained, these attacks have been shown to be effective to significantly degrade the performance of the system. In this paper we propose an efficient algorithm to perform optimal label flipping poisoning attacks and a mechanism to detect and relabel suspicious data points, mitigating the effect of such poisoning attacks.\footnote{Paper accepted at the Nemesis'18 Workshop on Recent Advances in Adversarial Machine Learning (co-located with ECML/PKDD 2018).}

\end{abstract}
\section{Introduction}
Many modern services and applications rely on data-driven approaches that use machine learning technologies to extract valuable information from the data received, provide advantages to the users, and allow the automation of many processes. However, machine learning systems are vulnerable and attackers can gain a significant advantage by compromising the learning algorithms. Thus, attackers can learn the blind spots and the weaknesses of the algorithm to manipulate samples at test time to evade detection or inject malicious data into the training set to poison the learning algorithm \cite{huang}. These attacks have already been reported in the wild against antivirus engines, spam filters, and systems aimed to detect fake profiles or news in social networks. 

Poisoning attacks are considered one of the most relevant and emerging security threats for data-driven technologies \cite{joseph}, especially in cases where the data is untrusted, as for example in IoT environments, sensor networks, applications that rely on the collection of users' data or where the labelling is crowdsourced from a set of untrusted annotators. Related work in adversarial machine learning has shown the effectiveness of optimal poisoning attacks to degrade the performance of popular machine learning classifiers --including Support Vector Machines (SVMs) \cite{biggioSVM}, embedded feature selection methods \cite{xiao}, neural networks and deep learning systems \cite{luis}-- by compromising a small fraction of the training dataset. Previous attacks assume that the attacker can manipulate both, the features and the labels of the poisoning points. For some applications this is not possible, and the attacker's capabilities are constrained to the manipulation of the labels. These are known as \emph{label flipping attacks}. Even if these attacks are more constrained, they are still capable of degrading significantly the performance of learning algorithms, including deep learning \cite{zhang}. 

Few \emph{general} defensive mechanisms have been proposed against poisoning attacks in the context of classification problems. For example, in \cite{nelson} the authors propose an algorithm that evaluates the impact of each training sample in the performance of the learning algorithms. Although this can be effective in some cases, the algorithm does not scale well for large datasets. In \cite{andrea}, an outlier detection scheme is proposed to identify and remove suspicious samples. Although the defensive algorithm is capable of successfully mitigating the effect of optimal poisoning attacks, its performance is limited to defend against label flipping attacks. Other more algorithm-dependent alternatives are described in Sect.~2.

In this paper we first propose an algorithm to perform label flipping poisoning attacks. The optimal formulation of the problem for the attacker is computationally intractable. We have developed an heuristic that allows to craft efficient label flipping attacks at a reduced computational cost. On the other hand, we also propose a defensive mechanism to mitigate the effect of label flipping attacks with label sanitization. We have developed an algorithm based on \emph{k}-Nearest-Neighbours ($k$-NN) to detect malicious samples or data points that have a negative impact on the performance of machine learning classifiers. We empirically show the effectiveness of our algorithm to mitigate the effect of label flipping attacks on a linear classifier for 3 real datasets. 

The rest of the paper is organised as follows: in Sect.~2 we describe the related work. In Sect.~3 we introduce a novel algorithm to perform optimal label flipping poisoning attacks. In Sect.~4 we present our defensive algorithm to mitigate the effect of label flipping attacks by identifying and relabelling suspicious samples. In Sect.~5 we show our experimental evaluation on real datasets assessing the validity of our proposed defence. Sect.~6 concludes the paper. 
\section{Related Work}
Optimal poisoning attacks against machine learning classifiers can be formulated as a bi-level optimization problem where the attacker aims to inject malicious points into the training set that maximize some objective function (e.g. increase the overall test classification error) while, at the same time, the defender learns the parameters of the algorithm by minimizing some loss function evaluated on the tainted dataset. This strategy has been proposed against popular binary classification algorithms such as SVMs \cite{biggioSVM}, logistic regression \cite{mei}, and embedded feature selection \cite{xiao}. An extension to multi-class classifiers was proposed in \cite{luis}, where the authors also devised an efficient algorithm to compute the poisoning points through back-gradient optimization, which allows to poison a broader range of learning algorithms, including neural networks and deep learning systems. An approximation to optimal poisoning attacks was proposed in \cite{koh} where the authors provide a mechanism to detect the most influential training points. The authors in \cite{zhang} showed that deep networks are vulnerable to (random) label noise. In \cite{biggioLabel}, a more advanced label flipping poisoning attack strategy is proposed against two-class SVMs, where the attacker selects the subset of training points that maximizes the error, evaluated on a separate validation set, when their labels are flipped. Label flipping attacks have been deeply investigated also by the computational learning theory community. In this community, the case where attacker can alter a fraction of the training data labels is often referred to as \emph{adversarial classification noise} and the goal is to design polynomial-time algorithms for PAC learning. However, solutions to this problem have only been found in very restricted distributional settings, e.g. isotropic log-concave marginal distributions realized by a linear model passing through the origin \cite{awasthi2017power}. However, in practice, the marginal data distribution is unknown and realizability is often violated, making these algorithms not very appealing for practical applications. Stronger results releasing realizability are known only for the case when even more restrictive assumptions on the attacker model are made \cite{awasthi2015efficient}.

Defences against optimal poisoning attacks typically consist either in identifying malicious examples and discarding them from the training data \cite{andrea} or they require to sove some robust optimization problem \cite{feng2014robust, DBLP:journals/corr/abs-1804-00308}. For example, the authors in \cite{andrea} propose the adoption of outlier detection algorithms to remove the poisoning data points from the training set before training. A \emph{white-box} model, where the attacker is aware of the outlier preprocessing, is considered in \cite{steinhardt}. In this latter work authors approximate a data-dependent upper bound on the performance of the learner under data poisoning with an online learning algorithm, assuming that some data sanitization is performed before training. Although the experimental evaluation supports the validity of this approach to mitigate optimal poisoning attacks, the capabilities of the algorithm to reduce the effect of more constrained attack strategies is limited. \cite{feng2014robust} proposes a robust version of logistic regression which comes with strong statistical guarantees, but it requires distributional assumption on the data generating distribution and assumes the poisoning level to be known. Instead, the defensive algorithm in \cite{DBLP:journals/corr/abs-1804-00308} iteratively trains the model while trimming the training data based on the current error estimates. However, \cite{DBLP:journals/corr/abs-1804-00308} focuses in the context of linear regression and the proposed defence requires to estimate a priori the number of poisoning data points. 

Specific mechanisms to defend against label flipping are described in \cite{nelson, koh}.  The authors in \cite{nelson} propose to measure the impact of each training example on the classifier's performance to detect poisoning points. Examples that affect negatively the performance are then discarded. Although effective in some cases, the algorithm scales poorly with the number of samples. Following the same spirit, a more scalable approach is proposed in \cite{koh} through the use of influence functions, where the algorithm aims to identify the impact of the training examples on the training cost function without retraining the model. 

\section{Label Flipping Attacks}
In a poisoning attack, the adversary injects malicious examples in training dataset to influence the behaviour of the learning algorithm according to some arbitrary goal defined by the attacker. Typically, adversarial training examples are designed to maximize the error of the learned classifier. In line with most of the related work, in this paper, we only consider binary classification problems. We restrict our analysis to \emph{worst-case} scenarios, where the attacker has perfect knowledge about the learning algorithm, the loss function that the defender is optimizing, the training data, and the set of features used by the learning algorithm. Additionally we assume that the attacker has access to a separate validation set, drawn from the same data distribution than the defender's training and test sets. Although unrealistic for practical scenarios, these assumptions allows us to provide worst-case analysis of the performance and the robustness of the learning algorithm when is under attack. This is especially useful for applications that require certain levels of assurance on the performance of the system.

We consider the problem of learning a binary linear classifier over a domain $\X \subseteq \R^d$ with labels in $\Y = \{-1, +1\}$. We assume that the classifiers are parametrized by $\vec{w} \in \R^d$, so that the output of the classifier is given by $h_{\vec{w}}(\vec{x}) = \sign( \vec{w}^\top \vec{x})$. We assume the learner to have access to an i.i.d. training dataset $S=\{(\vec{x}_i, y_i)\}_{i=1}^m$ drawn from an unknown distribution $\D$ over $\X \times \Y$. 


In a label flipping attack, the attacker's goal is to find a subset of $p$ examples in $S$ such that when their labels are flipped, some arbitrary objective function for the attacker is maximized. For the sake of simplicity, we assume that the objective of the attacker is to maximize the loss function, $\ell(\vec{w}, (\vec{x}_j, y_j))$, evaluated on a separate validation dataset $S_V = \{(\vec{x}_j, y_j)\}_{j=1}^n$. Then, let $\vec{u} \in \{0, 1\}^m$ with $\|\vec{u}\|_0 = p$ and let $S_p = \{P_i\}_{i =1}^m$ a set of examples defined such that: $P_i = (\vec{x}_i, y_i)$ if $\vec{u}(i) = 0$, and $P_i = (\vec{x}_i, -y_i)$ otherwise. Thus, $\vec{u}$ is an indicator vector to specify the samples whose labels are flipped and $S_p  = \{(\vec{x}'_i, y'_i)\}_{i=1}^m$ denotes the training dataset after those label flips. We can formulate the optimal label flipping attack strategy as the following bi-level optimization problem:
\vspace{-0.2cm}
\begin{equation}
\begin{aligned}
\vec{u}^* \in  \underset{\vec{u} \in \{0, 1\}^m, \|\vec{u}\|_0 = p}{\argmax} \quad & \frac{1}{n} \sum_{j = 1}^n \ell(\vec{w}, (\vec{x}_j, y_j)) \\
\text{s.t.}\quad & \vec{w} = {\cal A}_{\ell} (S_p) \\
\end{aligned}
\label{eq:lf}
\end{equation} where the parameters $\vec{w}$ are the result of a learning algorithm ${\cal A}_{\ell}$ that aims to optimize a loss function $\ell$ on the poisoned training set $S_p$.\footnote{For simplicity we assume that the attacker aims to maximize the average loss on a separate validation dataset.} Solving the bi-level optimization problem in (\ref{alg:lf}) is intractable, i.e. it requires a combinatorial search amongst all the possible subsets of $p$ samples in $S$ whose labels are flipped. To sidestep this difficulty in \Cref{alg:lf} we propose a heuristic to provide a (possibly) suboptimal but tractable solution to Problem (\ref{alg:lf}). Thus, our proposed algorithm greedily selects the examples to be flipped based on their impact on the validation objective function the attacker aims to maximize. At the beginning we initialize $\vec{u} = \vec{0}$, $S_p = S$, $I = [1,\ldots,m]$, where $I$ is the search space, described as a vector containing all possible indices in the training set. Then, at each iteration the algorithm selects from $S(I)$ the best sample to flip, i.e. the sample that,  when flipped, maximizes the error on the validation set, $e(j)$, given that the classifier is trained in the tainted dataset $S'$ (which contains the label flips from previous iterations). Then, the index of this sample, $i_j$, is removed from $I$, the $i_j$-th element of $\vec{u}$ is set to one, and $S_p$ is updated accordingly to the new value of $\vec{u}$.

\RestyleAlgo{boxruled}
\begin{algorithm}
\DontPrintSemicolon
\caption{Label Flipping Attack (LFA) \label{alg:lf}}
\textbf{Input:} training set $S = \{(\vec{x}_i, y_i)\}_{i=1}^m$, validation set $S_V = \{(\vec{x}_j, y_j)\}_{j=1}^n$, \# of examples to flip $p$. \\
\textbf{Initialize:} $u = \vec{0}, S_p = S, I = [m]$ \\
\For{$k \leftarrow 1$ \KwTo $p$}{
\For{$j \leftarrow 1$ \KwTo $|I|$}{
	$S' = S_p$, $S'_{I_j} = (\vec{x}_{I_j}, -y_{I_j})$ \\
	$\vec{w}^* \in {\argmin}_{\vec{w}} \frac{1}{m} \sum_{i = 1}^m \ell(\vec{w}, (\vec{x}'_i, y'_i))$ \\
	$e(j) = \frac{1}{n} \sum_{j = 1}^n \ell(\vec{w}, (\vec{x}_j, y_j))$ \\
}
$i_k = \argmax_{i \in I} e(i)$ \\
$u_{i_k} = 1, I = I/\{i_k\}$ \\
$S_p = S(\vec{u})$ \\
}
\textbf{Output:} poisoned training set $S_p$, flips $\vec{u}$
\end{algorithm}

\vspace{-0.2cm}
\section{Defence against Label Flipping Attacks}
We can expect aggressive label flipping strategies, such as the one described in Sect.~3, to flip the labels of points that are far from the decision boundary to maximize the impact of the attack. Then, many of these poisoning points will be far from the genuine points with the same label, and then, they can be considered as outliers.

To mitigate the effect of label flipping attacks we propose a mechanism to relabel points that are suspicious to be malicious. The algorithm uses $k$-NN to assign the label to each instance in the training set. The goal is to enforce label homogeneity between instances that are close, especially in regions that are far from the decision boundary. The procedure is described in \Cref{alg:def}. Thus, for each sample in the (possibly tainted) training set we find its $k$ nearest neighbours, $S_{k_i}$ using the euclidean distance.\footnote{Any other distance, such as the Hamming distance, can be applied, depending on the set of features used.} Then, if the fraction of data points in $S_{k_i}$ with the most common label in $S_{k_i}$ --denoted as $\conf(S_{k_i})$-- is equal or greater than a given threshold $\eta$, with $0.5 \leq \eta \leq 1$, the corresponding training sample is relabelled with the most common label in $S_{k_i}$. This can be expressed as $\mod(S_{k_i})$, the mode of the sample labels in $S_{k_i}$. The algorithm can also be repeated several times until no training samples are relabelled.

\vspace{-0.2cm}
\RestyleAlgo{boxruled}
\LinesNumbered
\begin{algorithm}
\DontPrintSemicolon
\caption{kNN-based Defence\label{alg:def}}
\textbf{Parameters:} $k$, $\eta$. \\
\textbf{Input:} training set $S = \{(\vec{x}_i, y_i)\}_{i=1}^m$. \\
\For{$i \leftarrow 1$ \KwTo $m$}{
$S_{k_i} = k$-NN$(S_{/i})$ \\
\lIf{$(\conf(S_{k_i}) \ge \eta)$}{$y_i' = \mod(S_{k_i})$}
\lElse{$y_i' = y_i$}
}
\textbf{Output:} $S$
\end{algorithm}
\vspace{-0.3cm}

Poisoning points that are far from the decision boundary are likely to be relabelled, mitigating their malicious effect on the performance of the classifier. Although the algorithm can also relabel genuine points, for example in regions where the two classes overlap (especially for values of $\eta$ close to $0.5$), we can expect a similar fraction of genuine samples relabelled in the two classes, so the label noise introduced by \Cref{alg:def} should be similar for the two classes. Then, the performance of the classifier should not be significantly affected by the application of our relabelling mechanism. Note that the algorithm is also applicable to multi-class classification problems, although in our experimental evaluation in Sect. 5 we only consider binary classification. 

\section{Experiments}
We evaluated the performance of our label flipping attack and the proposed defence on 3 real datasets from UCI repository:\footnote{\url{https://archive.ics.uci.edu/ml/datasets.html}} \emph{BreastCancer}, \emph{MNIST}, and \emph{Spambase}, which are common benchmarks for classification tasks. The characteristics of the datasets are described in \Cref{tab:datasets}. Similar to \cite{biggioSVM,luis}, for \emph{MNIST}, a multi-class problem for handwritten digits recognition, we transformed the problem into a binary classification task, aiming at recognising digits 1 and 7. As classifier, we used a linear classifier that aims to minimize the expected \emph{hinge loss}, $\ell(\vec{w}, (\vec{x}, y)) = \max\{0, 1 - y ( \vec{w}^\top \vec{x})\}$. We learned the parameters $\vec{w}$ with stochastic gradient descent.

\begin{table}
\begin{center}
\begin{tabular}{|c|c|c|c|}
\hline
\textbf{Name} & \textbf{\# Features} & \textbf{\# Examples} & \textbf{\# +/-} \\
\hline
BreastCancer & 30 & 569 & 212/357\\
MNIST (1 vs 7) & 784 & 13,007 & 6,742/6,265\\
SpamBase & 54 & 4,100 & 1,657/2,443 \\
\hline
\end{tabular}
\end{center}
\caption{Summary of the datasets used in the experiments. The rightmost column reports the number of positive and negative examples.}
\label{tab:datasets}
\end{table}

\begin{figure*}
\begin{center}
\subfloat[][BreastCancer]{\includegraphics[width=\columnwidth]{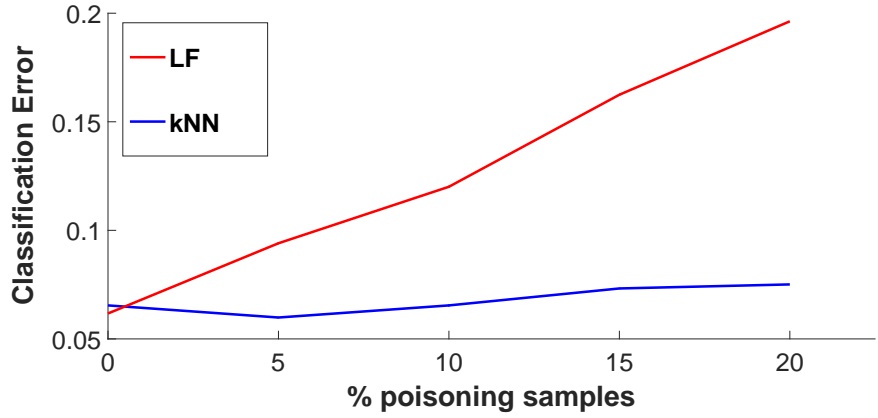}} \\
\subfloat[][MNIST]{\includegraphics[width=\columnwidth]{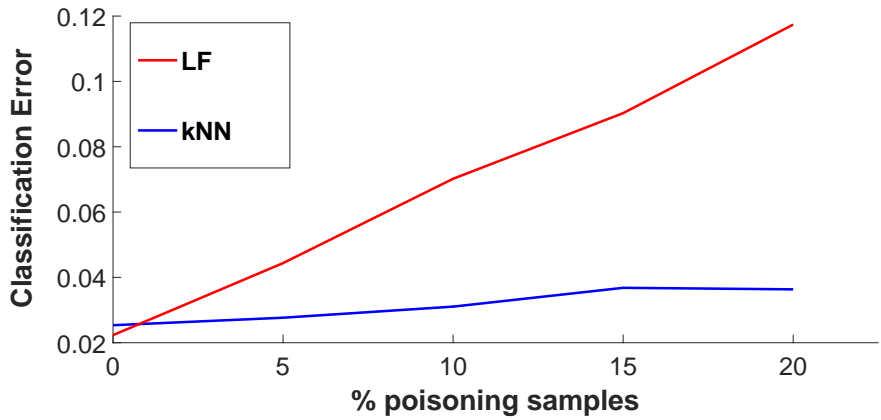}} \\
\subfloat[][(Spambase)]{\includegraphics[width=\columnwidth]{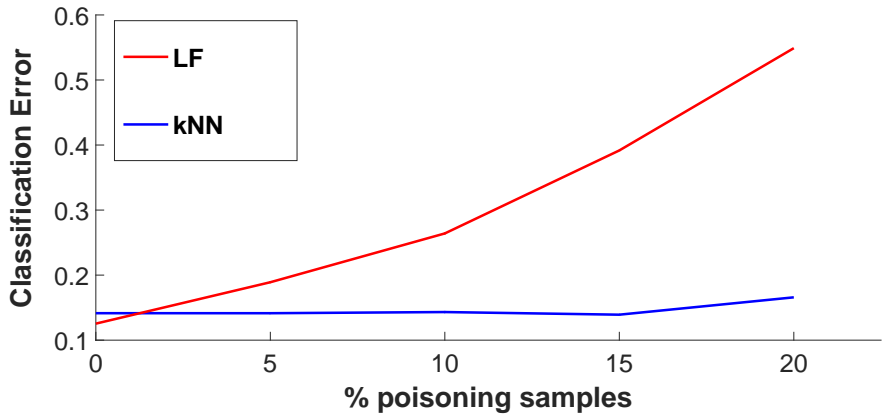}}
\\
\end{center}
\caption{Average classification error as a function of the percentage of poisoning points using the label flipping attack in \Cref{alg:lf}. Red line depicts the error when no defence is applied. Blue line shows the performance of the classifier after applying \Cref{alg:def}.}
\label{fig:resultsEx1}
\end{figure*}

In our first experiment we evaluated the effectiveness of the label flipping attack described in \Cref{alg:lf} to poison a linear classifier. We also assessed the performance of our defensive strategy in \Cref{alg:def} to mitigate the effect of this attack. For each dataset we created 10 random splits with $100$ points for training, $100$ for validation, and the rest for testing. For the learning algorithm we set the learning rate to $0.01$ and the number of epochs to $100$. For the defensive algorithm, we set the confidence parameter $\eta$ to $0.5$ and selected the number of neighbours $k$ according to the performance of the algorithm evaluated in the validation dataset. We assume that the attacker has not access to the validation data, so it cannot be poisoned. In practice, this requires the defender to have a small trusted validation dataset, which is reasonable for many applications. Note that typical scenarios of poisoning happen when retraining the machine learning system using data collected in the wild, but small fractions of data points can be curated before the system is deployed. From the experimental results in \Cref{fig:resultsEx1} we observe the effectiveness of the label flipping attack to degrade the performance of the classifier in the 3 datasets (when no defence is applied). Thus, after 20\% of poisoning, the average classification error increases by a factor of $2.8$, $6.0$, and $4.5$ respectively for \emph{BreastCancer}, \emph{MNIST}, and \emph{Spambase}. In \Cref{fig:resultsEx1} we also show that our defensive technique effectively mitigates the effect of the attack: The performance with 20\% of poisoning points is similar to the performance on the clean dataset on \emph{BreastCancer} and \emph{Spambase}, and we only appreciate a very slight degradation of the performance on \emph{MNIST}. When no attack is performed, we observe that our defensive strategy slightly degrades the performance of the classifier (compared to the case where no defence is applied). This can be due to the label noise introduced by the algorithm, which can relabel some genuine data points. However, this small loss in performance can be affordable for the sake of a more secure machine learning system. 

\begin{figure}
\begin{center}
\includegraphics[width=0.8\columnwidth]{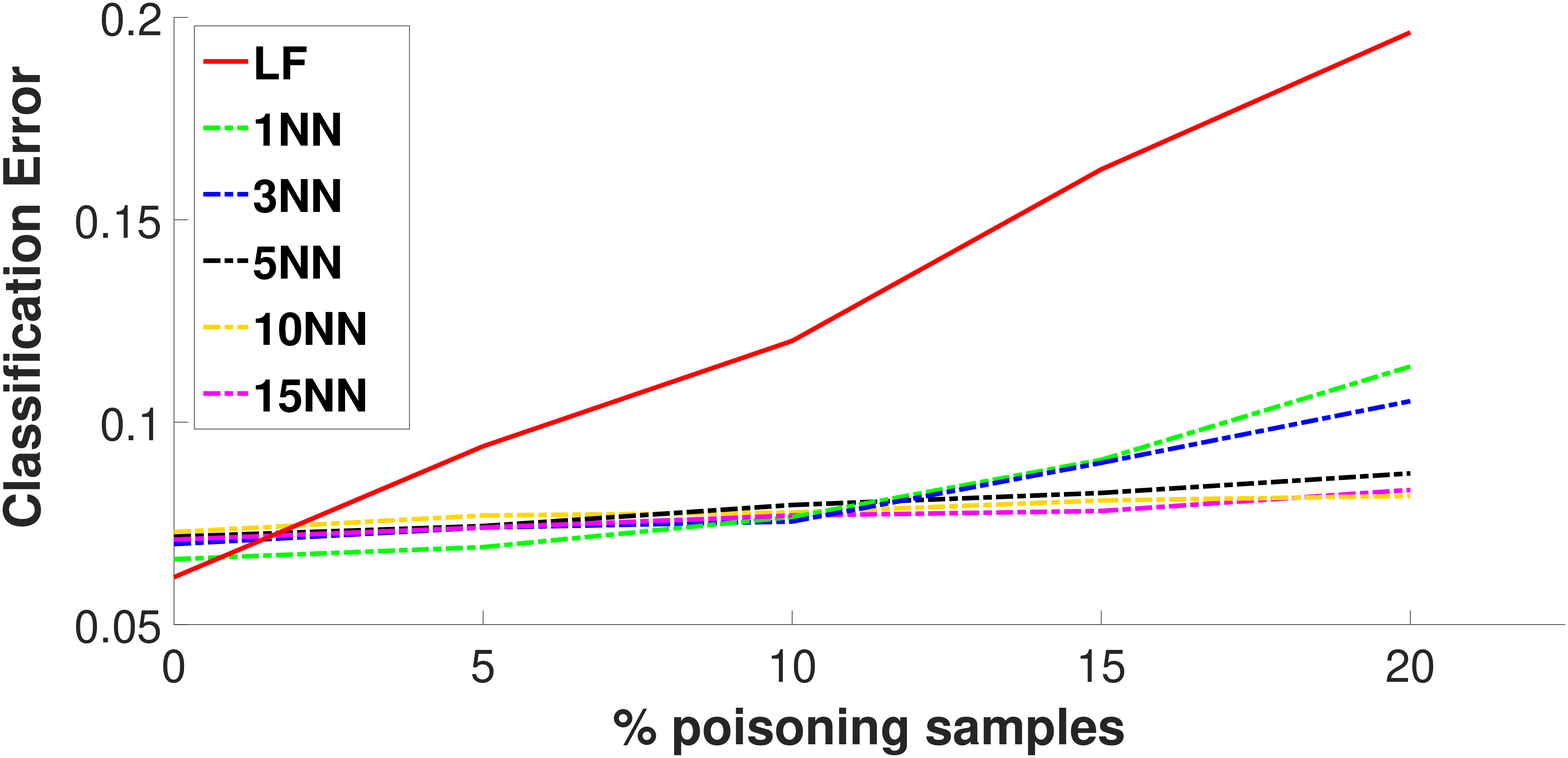} \\
(a) Sensitivity w.r.t. $k$ \\
\includegraphics[width=0.8\columnwidth]{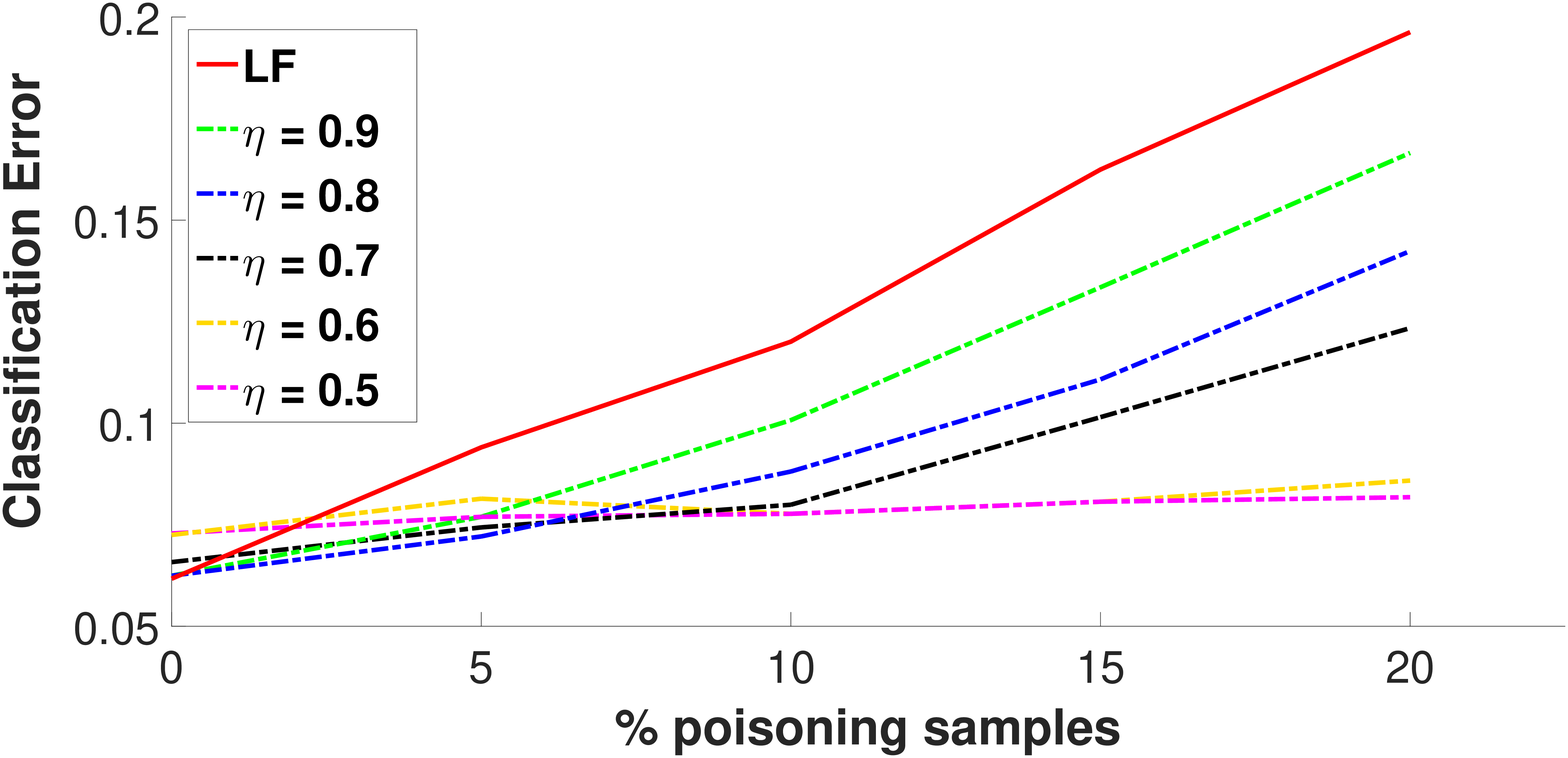} \\
(b) Sensitivity w.r.t. $\eta$
\\
\end{center}
\caption{Average test classification error as a function of the percentage of poisoning points. (a) Performance of the defensive algorithm for different values of $k$, with $\eta = 0.5$. (b) Performance for different values of $\eta$ for $k=10$. Solid red line depicts the baseline, when no defence is applied.}
\label{fig:resultsEx2}
\end{figure}

In \Cref{fig:resultsEx2} we show the sensitivity of the parameters $k$ and $\eta$ in \Cref{alg:def}. We report the average test classification error on \emph{BreastCancer} dataset for different configurations of our defensive strategy. In \Cref{fig:resultsEx2}.(a) we show the sensitivity of the algorithm to the number of neighbours $k$, setting the value of $\eta$ to $0.5$. We observe that for bigger values of $k$ the algorithm exhibits a better performance when the fraction of poisoning points is large, and the degradation on the performance is more graceful as the number of poisoning points increases. However, for smaller fractions of poisoning points or when no attack is performed, smaller values of $k$ show a slightly better classification error. In \Cref{fig:resultsEx2}.(b) we observe that \Cref{alg:def} is more sensitive to the confidence threshold $\eta$. Thus, for bigger values of $\eta$ the defence is less effective to mitigate the label flipping attacks, since we can expect less points to be relabelled. Then, small values of $\eta$ show a more graceful degradation with the fraction of poisoning points, although the performance when no attack is present is slightly worse.

\section{Conclusion}
In this paper we propose a label flipping poisoning attack strategy that is effective to compromise machine learning classifiers. We also propose a defence mechanism based on $k$-NN to achieve label sanitization, aiming to detect malicious poisoning points. We empirically showed the significant degradation of the performance produced by the proposed attack on linear classifiers as well as the effectiveness of the proposed defence to successfully mitigate the effect of such label flipping attacks. Future work will include the investigation of similar defensive strategies for less aggressive attacks, where the attacker considers detectability constraints. Similar to \cite{vittorio} we will also consider cases where the attack points collude towards the same objective, where more advanced techniques are required to detect malicious points and defend against these attacks.

%
%
%
%
\bibliographystyle{splncs04}
\bibliography{biblio}

\end{document}